\documentclass[onecolumn]{IEEEconf} 
\usepackage{nips12submit_e,times}
\usepackage[latin1]{inputenc}
\usepackage{amsmath}
\usepackage{amsfonts}
\usepackage{amssymb}
\usepackage{color}
\usepackage{graphicx}
\usepackage{bm}
\usepackage{algorithm,algorithmic}
\usepackage{multicol,multirow}
\usepackage{subfigure}
\usepackage{enumerate}

\usepackage{soul}
  \usepackage{color}
\definecolor{lightblue}{rgb}{.5,.85,1}
\definecolor{lightred}{rgb}{1,.4,.3}
\definecolor{lightorange}{rgb}{1,.7,.4}
\definecolor{yellow}{rgb}{1,1,0}

\providecommand{\rem}[1]{}%
\providecommand{\com}[1]{}

\DeclareMathOperator{\EV}{\mathrm{E}}  
\newtheorem{thm}{Theorem}[section]

\title{Rate-Distortion Auto-Encoders}

\author{
Luis G. {Sanchez Giraldo}
\\
 Dept. of Electrical and Computer Engineering\\
 University of Florida\\
 Gainesville, Florida, USA\\
\texttt{sanchez@cnel.ufl.edu} \\
\And
Jose C. Principe \\
 Dept. of Electrical and Computer Engineering\\
 University of Florida\\
 Gainesville, Florida, USA\\
\texttt{principe@cnel.ufl.edu} \\
}

\nipsfinalcopy 

\begin{document}
\maketitle

\begin{abstract}
A rekindled the interest in auto-encoder algorithms has been spurred by recent work on deep learning. Current efforts have been directed towards effective training of auto-encoder architectures with a large number of coding units. Here, we propose a learning algorithm for auto-encoders based on a rate-distortion objective that minimizes the mutual information between the inputs and the outputs of the auto-encoder subject to a fidelity constraint. The goal is to learn a representation that is minimally committed to the input data, but that is rich enough to reconstruct the inputs up to certain level of distortion. Minimizing the mutual information acts as a regularization term whereas the fidelity constraint can be understood as a risk functional in the conventional statistical learning setting. The proposed algorithm uses a recently introduced measure of entropy based on infinitely divisible matrices that avoids the plug in estimation of densities. Experiments using over-complete bases show that the rate-distortion auto-encoders can learn a regularized input-output mapping in an implicit manner.  
\end{abstract}

\section{Introduction}\label{sec:Introduction}
Auto-encoders are unsupervised learning algorithms that capture the structure in data by finding an internal representation of the input patterns (encoding) from which they can be reconstructed at least approximately. By learning a transformation $G$ (encoding) from the input $x \in \mathcal{X}$ to $z = G(x) \in \mathcal{Z}$ the auto-encoder tries to capture the structure of the input. To guarantee the transformation $G$ preserves the information about $x$, a decoder $\tilde{G}^{-1}$  is also learned along way, such that a measure of fidelity $\EV[D(X,\tilde{X})]$ in the reconstruction is optimized. Perhaps, the most conventional class of auto-encoders are the bottle-neck networks that transform the input data $x \in \mathcal{X} \subseteq \mathbb{R}^d$ to a lower dimensional space $\mathcal{Z} \subseteq \mathcal{R}^p$, and then transform it back to the input space $\mathcal{X}$. It is expected that the lower dimensional space forces the encoder to capture the relations between the variables in the input space. While this is clear when one is restricted to linear mappings, the problem becomes less well-understood if nonlinear mappings are allowed. \\
Recent progress in deep learning has reignited the interest on developing procedures to train auto encoders. Nevertheless, research efforts are now focused on how to effectively train auto-encoder architectures with a large number of encoding units, possibly larger that the number of inputs. In this case, it is necessary to develop algorithms that can avoid trivial solutions. Approaches like sparse encoding, which can be applied to over-complete scenarios, $\mathrm{dim}(\mathcal{X}) < \mathrm{dim}(\mathcal{Z})$, use the concept of effective dimensionality. A sparsity constraint induces an active set of variables with an expected $L^0$ norm smaller than the input space dimensionality. In general, sparse coding procedures require solving an optimization problem at inference time since the encoding mapping is not explicit. However, it has been shown that efficient inference can be made possible by training  a nonlinear feed-forward network to mimic the output of a sparse encoding algorithm \cite{MRanzato06}. \\
Another recent approach focus on the idea of robustness. De-noising auto-encoders \cite{PVincent08} learn a mapping by simply minimizing the error between the reconstruction of the auto-encoder when a noisy version of the input is feed to the mapper. Let $X$ be the random variable representing  the input, and $\hat{X}$ be the noisy version of it. The goal of the de-noising auto-encoder is to learn a map $f(\cdot) = \tilde{G}^{-1}(G(\cdot))$ such that $\EV[D(f(\hat{X}), X)]$ is minimized. Contractive auto-encoders \cite{SRifai11A, SRifai11B} are another way of enforcing robustness on the learned encoder by penalizing the sensitivity of the representation $G(x)$ to the input $x$. This is achieved by minimizing the expected Frobenius norm of the Jacobian $\EV[J_{G}(X)]$ of $G$ while maximizing fidelity of the reconstruction.

Here, we adopt an information theoretic view of the problem. In all the above examples, implicitly or explicitly, the concept of compression plays a fundamental role. The idea of rate-distortion has been explored for nondeterministic mappings by \cite{WBialek04}. In this case the structure of the data is represented by a set of $N$ points located along a manifold. The algorithm produces random map $p(\mu|x)$ from $\mathcal{X}$ to the embedding of a  manifold denoted by $\mathcal{M}$. The manifold is assumed to be a less faithful description of the data. The objective is to minimize a trade-off between fidelity and information content. The information content correspond to the mutual information between the data and the manifold representation: 
\begin{equation}\label{eq:mutual_information_data_manifold}
I(X;\mathcal{M}) = \int\limits_{x \in \mathcal{X}}\int\limits_{\mu \in \mathcal{M}}P(x, \mu)\log{\frac{P(x, \mu)}{p(x), P_{\mathcal{M}}(\mu)}} \mathrm{d}x
\end{equation}  
The locations of the points are adjusted according to the distortion measure. A Boltzmann distribution containing the fidelity measure arises as solution to the above objective. The update equations for an algorithm based on the above trade off correspond to an iterated projection between convex sets. Note however, that a solution of this form arises from the considerations of a random mapping from $\mathcal{X}$ to the embedded manifold. 

Here, we propose an algorithm that incorporates the idea to learn a parametrized mapping, encoder and decoder function, that optimizes a trade off between compression and fidelity in the reconstruction. The intuition behind this procedure is that the auto-encoder should avoid trivial solutions if it is forced to map the input data to a compressed version of it. 

\section{The rate-distortion function}
 For the transmission of information through a channel, it was showed by Shannon that the noise and bandwidth inherent to the channel determine the rate at which information can be transfered, and that any attempt to transfer information above the limit imposed by the channel will incur in some error in the recovered message. This result known as the channel capacity problem is concerned with achieving zero error in the transmission. A complementary problem is to determine what is the minimum possible rate at which information can be transfered such that the error in the recovered message does not surpasses a certain level. Answering this question gives rise to  the rate-distortion function. The information rate-distortion function $R(D)$ is defined as:
\begin{equation}\label{eq:Rate_distortion_function}
R(D) = \min\limits_{q(\hat{x}|x)}I(X;\hat{X}) \:\:\textrm{subject to}\:\:\EV[d(X,\hat{X})] \leq D
\end{equation}
where $d(\cdot,\cdot)$ is a distortion function that can be related to the loss incurred when the exact value of the input cannot be recovered from the output and $\EV[d(X,\hat{X})] = \sum_{x,\hat{x}} p(x)q(\hat{x}|x)d(x,\hat{x})$ . The above problem \eqref{eq:Rate_distortion_function} is a variational problem for which we search for $q(\hat{x}|x) \geq 0$ subject to the regularity condition:
\begin{equation}\label{eq:rate_distortion_regularity}
\sum\limits_{\hat{x} \in \hat{\mathcal{X}}}q(\hat{x}|x) = 1 \:\: \textrm{for all}\:x \in \mathcal{X} 
\end{equation}
This is a standard minimization problem of a convex function over a convex set \cite{TBerger} with a know numerical solution in the case of finite input/output alphabets. How could this function be utilized to find a representation of the input variables?
\subsection{Motivation Example: rate-distortion and PCA}
The relation between rate-distortion and PCA arises from the assumption of a Gaussian source. Consider the Gaussian random vector $X \in \mathbb{R}^d$ with zero mean and covariance $\bm{\Sigma}$. We can find similarity transformation $\mathbf{U}$ to a Gaussian random vector $Z \in \mathbb{R}^d$ with zero mean and diagonal covariance matrix $\bm{\Gamma}$, such that $\mathbf{U}^T \bm{\Sigma}\mathbf{U} = \bm{\Gamma}$. The transformation matrix $\mathbf{U}$ is unitary  which implies $\det{(\mathbf{U})} = 1$, and thus, $h(X) = h(Z)$ and the mapping is a bijection \footnote{Note, we are not using mutual information to relate $X$ and $Z$, cause in this case it is infinite}. The following theorem from \cite{TCover} is used on stating the result.
\begin{thm}\textbf{\emph{(Rate distortion for a parallel Gaussian source):}}
Let $X_i \sim \mathcal{N}(0,\gamma_i^ 2)$, $i=1, \dots, d,$ be independent Gaussian random variables, and the distortion measure $d(x,\hat{x}) = \sum_{i=1}^d(x_i - \hat{x}_i)^ 2$. Then the rate distortion function is given by
\begin{equation}
R(D) = \sum\limits_{i=1}^d\frac{1}{2}\log{\frac{\gamma_i}{D_i}}
\end{equation}  
where
\begin{equation}
D_i = \left\{\begin{array}{lr}
\lambda & \textrm{if}\: \lambda < \gamma_i^2 \\
\gamma_i^2 & \textrm{if}\: \lambda \geq \gamma_i^2, 
\end{array}\right.
\end{equation}
and $\lambda$ is chosen so that $\sum_{i=1}^dD_i =D$.
\end{thm}
Applying the result to the random vector $Z$ yields the conclusion that the rate for components with variance less than $\lambda$ will be zero, and thus, no effort in representing such components should be made. This is the way components are selected in PCA where a variance threshold is set and only those components above the threshold are retained. On the other hand, $\lambda$ plays the role of the observation noise in the generative model for PCA, but the main difference is that $\hat{Z}$ is not assumed to be $\mathcal{N}(0,\mathbf{I})$.
Note however that for the above conclusion, the similarity transformation was given in advance rather than derived from the rate-distortion objective. Our goal is to learn a map $f: \mathcal{X} \mapsto \mathcal{X}$ based on the principle of minimization of mutual information. The numerical procedure know as the Blahut-Arimoto algorithm provides a means to compute a stochastic map $q(\hat{x}|x)$ that achieves the minimum rate under a given distortion constraint. Nevertheless, this method was developed for finite input/output alphabets. We are interested in an approximation that uses a deterministic map $f(\cdot)$ on continuous input/output spaces.

\subsection{Rate-distortion as an objective for learning a map}
Shannon's definition of mutual information can be decomposed in terms of the marginal and conditional entropies as:
\begin{equation}
I(X;Y) = H(X) - H(X|Y)
\end{equation}
where $H(X)$ and $H(X|Y)$ are the marginal and conditional entropies of $X$ and $X$ given $Y$. Since the entropy of the input $X$ is constant, we only need to consider $H(X|Y)$ when dealing with maps from $X$ to $Y$. We can thus reformulate \eqref{eq:Rate_distortion_function} in terms of the conditional entropy as:
\begin{equation}\label{eq:Rate_distortion_function_cond_entropy}
\underset{f(X) = \hat{X}}{\mathrm{maximize}}\: H(X|\hat{X}) \:\:\textrm{subject to}\:\:\EV[d(X,\hat{X})] \leq D.
\end{equation}
Writing the Lagrangian of \eqref{eq:Rate_distortion_function_cond_entropy},
\begin{equation}\label{eq:Rate_distortion_function_cond_entropy_lagrangian}
\mathcal{L}(f,\gamma) = \EV[d(X,\hat{X})] - \frac{1}{\mu}H(X|\hat{X}), 
\end{equation} 
where $\mu$ determines the distortion level $D$, shows that the the objective can be understood as a regularized risk minimization problem. In \eqref{eq:Rate_distortion_function_cond_entropy_lagrangian}, $d(X,\hat{X})$ plays the role of the loss functional and $H(X|\hat{X})$ acts as a regularization parameter. However, this form of regularization is not explicit on the parameters of the function $f(\cdot)$. Informally, we can see that maximizing the conditional entropy $H(X|\hat{X}) = H(X;\hat{X}) - H(\hat{X})$ can have the effect of lowering the entropy of the output variable $\hat{X}$ and thus, we can think of the mapping $f$ as a contraction.  

\section{Rate-Distortion Auto-Encoder Algorithm}
On the previous section, we motivate the use of the rate-distortion objective to learn a map that is minimally committed in terms of retaining information about the inputs, yet able to provide a reconstruction with a desired level of distortion. However, solving the above objective function would require the distribution of $X$ to be known. Since we only have access to a set of \emph{i.i.d.} samples $\{\mathbf{x}_i\}_{i=1}^N$ drawn from $P_{X}(x)$, a suitable estimator of both conditional entropy, and expected distortion is required. For the expected distortion we employ the empirical estimator 
\begin{equation}\label{eq:empirical_expected_distortion}
D_{\textrm{emp}}(f) = \frac{1}{N}\sum\limits_{i=1}^N d(\mathbf{x}_i, f(\mathbf{x}_i)).
\end{equation}
For the conditional entropy term, we use an alternative definition of entropy introduced in \cite{LSanchez13}.   
\subsection{Matrix-Based Entropy Functional}
Let $\mathbf{X} = \{\mathbf{x}_i\}_{i=1}^N \subset \mathcal{X}$ and $\kappa:\mathcal{X}\times\mathcal{X} \mapsto \mathbb{R}$ be a real valued positive definite kernel that is also infinitely divisible. The Gram matrix $K$ obtained from evaluating a positive definite kernel $\kappa(\cdot, \cdot)$ on all pairs of exemplars, that is $(K)_{ij} = \kappa(\mathbf{x}_i, \mathbf{x}_j)$, can be employed to define a quantity with properties similar to those of an entropy, for which the probability distribution of $X$ does not need to be estimated.\\
A matrix-based analogue to Renyi's $\alpha$-entropy for a positive definite matrix $A$  of size $N\times N$, such that $\mathrm{tr}{(A)}  = 1$, is given by the functional:
\begin{equation}\label{eq:renyi_matrix_entropy}
S_{\alpha}(A) = \frac{1}{1-\alpha}\log_{2}{\left[\mathrm{tr}{(A^{\alpha})}\right]} = \frac{1}{1-\alpha}\log_{2}{\left[\sum\limits_{i=1}^N\lambda_i(A)^{\alpha}\right]},
\end{equation}
where $\lambda_i(A)$ denotes the $i$th eigenvalue of the $A$ matrix\footnote{All eigenvalues are considered along with their multiplicities}.The matrix-based entropy estimate from the sample $\mathbf{X}$ can be obtained by evaluating \eqref{eq:renyi_matrix_entropy} on following normalized version of $K$: 
\begin{equation}\label{eq:matrix_normalization}
A_{ij} = \frac{1}{N}\frac{K_{ij}}{\sqrt{K_{ii}K_{jj}}}.
\end{equation} 
Furthermore, based on the the product kernel, we can define the notion of joint-entropy as:
\begin{equation}\label{eq:hadamard_joint_entropy}
S_{\alpha}\left(\frac{A \circ B}{\mathrm{tr}(A \circ B)}\right),
\end{equation}
where $A\circ B$ denotes the Hadamard product between the matrices $A$ and $B$. For a set of pairs $\{(x_i, y_i)\}_{i=1}^n$ such that $x_i \in \mathcal{X}$ and $y_i \in \mathcal{Y}$, let $\kappa_{\mathcal{X}}$ and $\kappa_{\mathcal{Y}}$ be positive definite kernels defined on $\mathcal{X} \times \mathcal{X}$ and $\mathcal{Y} \times \mathcal{Y}$, respectively. If $A_{ij} = \kappa_{\mathcal{X}}(x_i, x_j)$ and $B_{ij} = \kappa_{\mathcal{Y}}(y_i, y_j)$, the Hadamard product $A \circ B$ is nothing but the Gram matrix using the product kernel $\kappa((x_i, y_i),(x_j, y_j)) = \kappa_{\mathcal{X}}(x_i, x_j)\cdot\kappa_{\mathcal{Y}}(y_i, y_j)$ \cite{CBerg}. It can also be shown that if the Gram matrices $A$ and $B$ are constructed using normalized infinitely divisible kernels (based on \eqref{eq:matrix_normalization}), such that $A_{ii} = B_{ii} = 1/n$, \eqref{eq:hadamard_joint_entropy} is never smaller than any of the individual entropies $S_\alpha(A)$ or $S_\alpha(B)$. This allows us to define a matrix notion of conditional entropy as follows:
\begin{equation}\label{eq:joint_entropy_gap}
S_{\alpha}(A|B) = S_{\alpha}\left(\frac{A \circ B}{\mathrm{tr}{(A\circ B)}}\right) - S_{\alpha}(B), 
\end{equation}
\subsection{Maximum Matrix-Based Conditional Entropy Auto-encoder}
Using the kernel matrix measure of entropy, we can formulate an objective function based on \eqref{eq:joint_entropy_gap}. For a set of points $\{\mathbf{x}_i\}_{i=1}^n$, where $\mathbf{x}_i \in \mathbb{R}^{d_x}$ we define a parametrized encoding mapping $\hat{\mathbf{z}}_i = G_{\mathbf{W},\mathbf{c}}(\mathbf{x}_i) = g(\mathbf{W}\mathbf{x}_i + \mathbf{c})$, as well as a decoding mapping $\tilde{\mathbf{x}}_i = \tilde{G}_{\mathbf{A},\mathbf{b}}^{-1}(\hat{\mathbf{z}}_i) = \mathbf{A}\hat{\mathbf{z}}_i + \mathbf{b}$. Notice that there is no imposed constraint on the mapping such as tied weights or explicit weight decay in our formulation. Let $\mathbf{K}_{X}$ and $\mathbf{K}_{\hat{X}}$ denote the normalized Gram matrices for $\{\mathbf{x}_i\}_{i=1}^n$ and $\{\hat{\mathbf{x}}_i\}_{i=1}^n$, respectively. Our goal is to find the set of pairs $\left(\mathbf{W}, \mathbf{c}\right)$ and $\left(\mathbf{A}, \mathbf{b}\right)$ that maximize the matrix-based conditional entropy of the inputs $\mathbf{x}_i$ given the reconstructions $\tilde{\mathbf{x}}_i = \tilde{G}_{\mathbf{A}}^{-1}(G_{\mathbf{W}}(\mathbf{x}_i))$. This can be posed as the following optimization problem:
\begin{equation}\label{eq:RD_autoencoder_optimization}
\begin{split}
\underset{(\mathbf{W}, \mathbf{c}, \mathbf{A}, \mathbf{b}) \in \bm{\Theta}}{\textrm{maximize}} & \:\:S_{\alpha}(\mathbf{K}_X \vert \mathbf{K}_{\tilde{X}})\\
\textrm{subject to} &\:\: \begin{array}{l}
\frac{1}{N}\sum\limits_{i=1}^N d(\mathbf{x}_i, \tilde{\mathbf{x}}_i) \leq D.
\end{array}
\end{split}
\end{equation}
Let $\mathbf{X} = (\mathbf{x}_1,\;\mathbf{x}_2,\cdots,\mathbf{x}_N)^{\mathrm{T}}$ denote the input data matrix, $\mathbf{Z} = \mathbf{X}\mathbf{W}^{\mathrm{T}} + \mathbf{1}_{N}\mathbf{c}^{\mathrm{T}}$ be the affine transform of the input , $\hat{\mathbf{Z}} = g(\mathbf{Z})$ denote the encoder output after the nonlinearity $g(\cdot)$, $\mathbf{G'} = g'(\mathbf{Z})$ be the derivatives of the encoder nonlinearity evaluated at $\mathbf{Z}$, and $\hat{\mathbf{X}} = \hat{\mathbf{Z}}\mathbf{A}^{\mathrm{T}} + \mathbf{1}_{N}\mathbf{b}^{\mathrm{T}}$ be the decoder output. The gradients of $S_{\alpha}(\mathbf{K}_{\hat{X}})$ and $S_{\alpha}(N\mathbf{K}_{X}\circ\mathbf{K}_{\hat{X}})$ with respect to $\mathbf{K}_{\hat{X}}$ are given by:
\begin{eqnarray}
\frac{\partial S_{\alpha}(\mathbf{K}_{\hat{X}})}{\partial \mathbf{K}_{\hat{X}}} & = & \frac{\alpha}{(1-\alpha)\mathrm{tr}(\mathbf{K}_{\hat{X}}^{\alpha})}\mathbf{U}\bm{\Lambda}^{\alpha-1} \mathbf{U}^{\mathrm{T}} \:\:\textrm{and}\\ 
\frac{\partial S_{\alpha}(N\mathbf{K}_{X}\circ\mathbf{K}_{\hat{X}})}{\partial \mathbf{K}_{\hat{X}}} & = &  \frac{\alpha}{(1-\alpha)\mathrm{tr}[(N\mathbf{K}_{X}\circ\mathbf{K}_{\hat{X}})^{\alpha}]}N\mathbf{K}_{X}\circ \left(\mathbf{V}\bm{\Gamma}^{\alpha-1} \mathbf{V}^{\mathrm{T}}\right),
\end{eqnarray}
where $\mathbf{U}\bm{\Lambda} \mathbf{U}^{\mathrm{T}}$  and $\mathbf{V}\bm{\Gamma} \mathbf{V}^{\mathrm{T}}$ are the eigenvalue decompositions of $\mathbf{K}_{\hat{X}}$ and $N\mathbf{K}_{X}\circ\mathbf{K}_{\hat{X}}$, respectively. In our work, we use the normalized Gaussian kernel, $\kappa(\mathbf{x}_i,\mathbf{x}_j)  = \frac{1}{n}\exp{\left(-\frac{\Vert\mathbf{x}_i-\mathbf{x}_j\Vert^2}{2\sigma_{x}^2}\right)}$, for both input and output points. The partial derivatives of the conditional entropy with respect to the parameters of the auto-encoder are given by:
\begin{eqnarray}
\label{eq:part_entropy_W}  \frac{\partial S_{\alpha}(\mathbf{K}_{\hat{X}})}{\partial \mathbf{A}} & = & -4(\mathbf{A}\hat{\mathbf{Z}}^{\mathrm{T}})\left[\mathcal{D}(\mathbf{P}) - \mathbf{P} \right]\hat{\mathbf{Z}}, \\
\label{eq:part_entropy_c}  \frac{\partial S_{\alpha}(\mathbf{K}_{\hat{X}})}{\partial \mathbf{b}} & = &  \mathbf{0}, \\
\label{eq:part_entropy_A}  \frac{\partial S_{\alpha}(\mathbf{K}_{\hat{X}})}{\partial \mathbf{W}} & = &  -4\left((\mathbf{A}^{\mathrm{T}}\mathbf{A}\hat{\mathbf{Z}}^{\mathrm{T}}\left[\mathcal{D}(\mathbf{P}) - \mathbf{P}\right])\circ \mathbf{G'}^{\mathrm{T}}\right)\mathbf{X}, \\ 
\label{eq:part_entropy_b}  \frac{\partial S_{\alpha}(\mathbf{K}_{\hat{X}})}{\partial \mathbf{c}} & = & -4\left((\mathbf{A}^{\mathrm{T}}\mathbf{A}\hat{\mathbf{Z}}^{\mathrm{T}}\left[\mathcal{D}(\mathbf{P}) - \mathbf{P}\right])\circ \mathbf{G'}^{\mathrm{T}}\right)\mathbf{1}_N,   
\end{eqnarray}
where $\mathbf{P} = \frac{\partial S_{\alpha}(\mathbf{K}_{\hat{X}})}{\partial \mathbf{K}_{\hat{X}}} \circ \frac{1}{2\sigma_{\hat{x}}^2}\mathbf{K}_{\hat{X}}$ and $\mathcal{D}(\mathbf{P}) = \mathrm{diag}(\mathbf{P}\mathbf{1}_N)$. The partial derivatives of the joint entropy term have the same form of the above, we only need to replace $\mathbf{P}$ by $\mathbf{Q} = \frac{\partial S_{\alpha}(N\mathbf{K}_{X}\circ\mathbf{K}_{\hat{X}})}{\partial \mathbf{K}_{\hat{X}}} \circ \frac{1}{2\sigma_{\hat{x}}^2}\mathbf{K}_{\hat{X}}$. From the above set of derivatives, \eqref{eq:part_entropy_W},\eqref{eq:part_entropy_A}, and \eqref{eq:part_entropy_b}, we can see that the term $\mathcal{D}(\mathbf{P}) - \mathbf{P}$ appears in all of them. Noticing that $\mathbf{P}$ is a positive definite and that indeed can be regarded as an affinity matrix, we can think of $\mathcal{D}(\mathbf{P}) - \mathbf{P}$ as the graph Laplacian, where $\mathbf{P}$ is the weight matrix for the set of edges and $\mathcal{D}(\mathbf{P})$ the degree of the graph. Notice also that $\mathbf{P}$ corresponds to a data dependent kernel since its values depend on the spectrum of $\mathbf{K}_{\hat{X}}$.\\
The derivatives with respect to the distortion measure $D_{\textrm{emp}}$ are given by:
\begin{eqnarray}
\label{eq:part_distortion_W}  \frac{\partial D_{\textrm{emp}}}{\partial \mathbf{A}} & = & -\frac{2}{N}(\mathbf{X}-\hat{\mathbf{X}})^{\mathrm{T}}\hat{\mathbf{Z}}, \\
\label{eq:part_distortion_c}  \frac{\partial D_{\textrm{emp}}}{\partial \mathbf{b}} & = & -\frac{2}{N}(\mathbf{X}-\hat{\mathbf{X}})^{\mathrm{T}}\mathbf{1}_N , \\
\label{eq:part_distortion_A}  \frac{\partial D_{\textrm{emp}}}{\partial \mathbf{W}} & = & -\frac{2}{N}\left[ \left( (\mathbf{X}-\hat{\mathbf{X}})\mathbf{A}^{\mathrm{T}} \right)\circ \mathbf{G'} \right]^{\mathrm{T}}\mathbf{X}, \\ 
\label{eq:part_distortion_b}  \frac{\partial D_{\textrm{emp}}}{\partial \mathbf{c}} & = & -\frac{2}{N}\left[ \left( (\mathbf{X}-\hat{\mathbf{X}})\mathbf{A}^{\mathrm{T}} \right)\circ \mathbf{G'} \right]^{\mathrm{T}}\mathbf{1}_N,   
\end{eqnarray}
The above derivatives can be employed to search for the parameters for instance using gradient ascent. In our work we implemented gradient ascent for the following Lagrangian:
\begin{equation}
\mathcal{L}(\mathbf{W},\mathbf{c},\mathbf{A},\mathbf{b}) = S_{\alpha}(N\mathbf{K}_{X}\circ\mathbf{K}_{\hat{X}}) - S_{\alpha}(\mathbf{K}_{\hat{X}}) - \mu \frac{1}{N}\sum\limits_{i=1}^N\Vert\mathbf{x}_i - \hat{\mathbf{x}}_i\Vert^2,
\end{equation}
where $\mu$ is fixed depending on the desired distortion level\footnote{Note that by fixing the value of $\mu$ we are implicitly enforcing an equality constraint on $D_{\textrm{emp}}$ rather than an inequality constraint since the positive multiplier represents an active constraint.}. 
\section{Experiments}
This section describe some of the experiments we have carried out with the rate-distortion auto-encoders.
We qualitatively illustrate the regularization property of the rate-distortion auto-encoder with a simple set of examples involving an over-complete representations. First,we show how the rate-distortion objective regularizes different types of units implicitly. Algorithm such as contractive auto-encoder rely on explicit calculation of the Jacobian, which depends on the type of units selected. Following these experiments, we show some resulting basis after learning on the MNIST dataset.  
\subsection{Synthetic Data}
\begin{figure}
\centering
\subfigure[Non regularized]{\includegraphics[width=4.6cm]{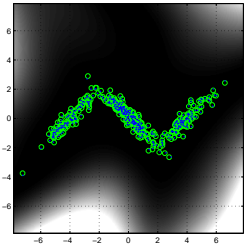}}
\subfigure[De-noising auto-encoder noise $\sigma = 0.5$]{\includegraphics[width=4.6cm]{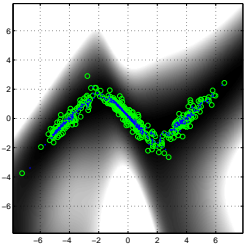}}
\subfigure[Rate-distortion auto-encoder]{\includegraphics[width=4.6cm]{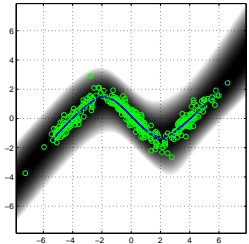}}
\caption{Energy $\Vert x - \hat{x} \Vert^2$ landscapes for different auto-encoder algorithms}\label{fig:RDAE_Gaussian_Mixture_Energy}
\end{figure}

\textbf{Gaussian distributed data:} The first example corresponds to a set of data point drawn from a bivariate Gaussian distribution with zero mean and covariance matrix 
\begin{equation}
\bm{\Sigma}_X = \left(
\begin{array}{cc}
1 & 0.95 \\
0.95 & 1
\end{array}
\right).
\end{equation}
We use a linear encoding, that is $g(x) = x$, that is also over-complete since the encoder project the $2$-dimensional data points to $10$ different directions. The conventional auto encoder would over fit being able to achieve zero reconstruction error, but it won't be able to implicitly retain what is thought to be the structure in the data. We also compare this output to the outputs of two nonlinear auto-encoders, one uses the $\mathrm{logsig}$ units $g(x) = 1/(1+\exp(-x))$, and the other a rectified linear units (ReLU) $g(x) = \max\{0,x\}$.
\begin{figure}[!t]
\centering
\subfigure[Linear $g(x) = x$]{\includegraphics[width=4.6cm]{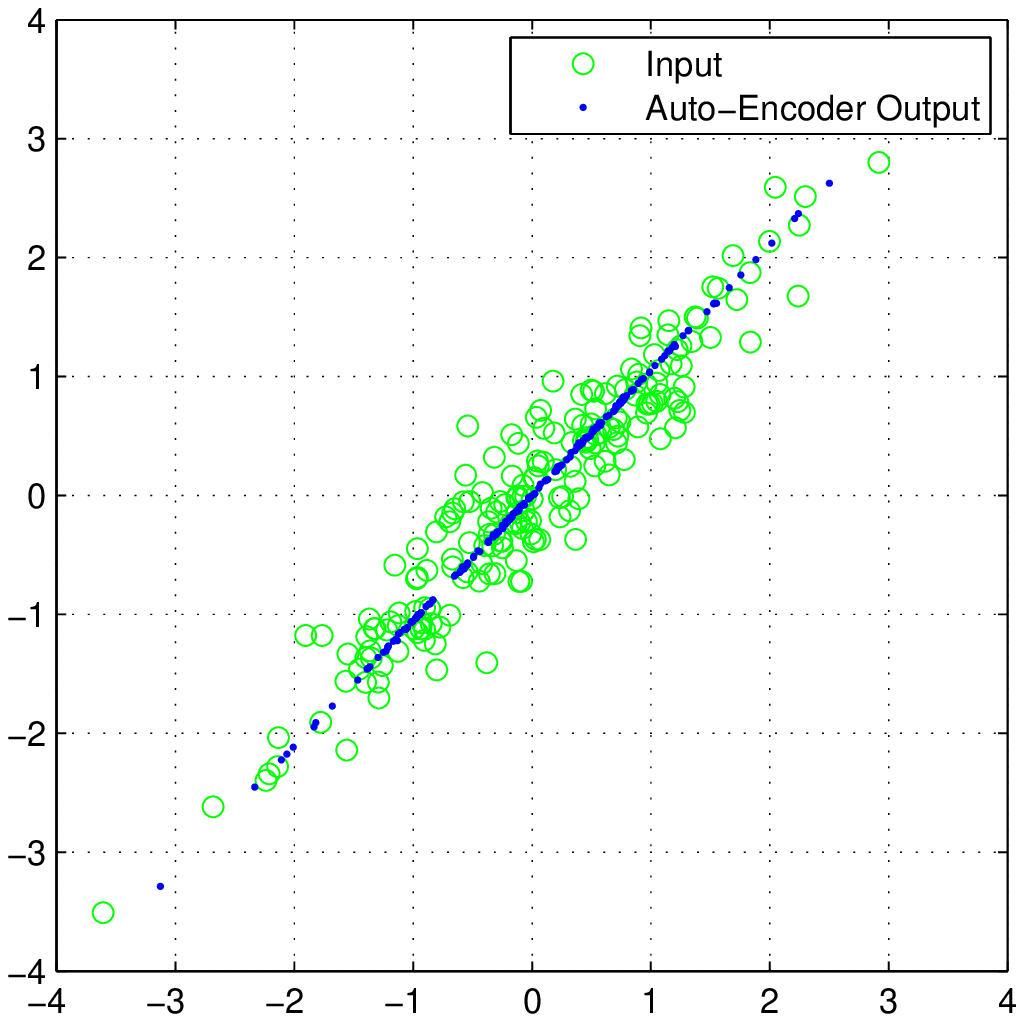}}
\subfigure[ReLU $g(x) = \max\{0,x\}$]{\includegraphics[width=4.6cm]{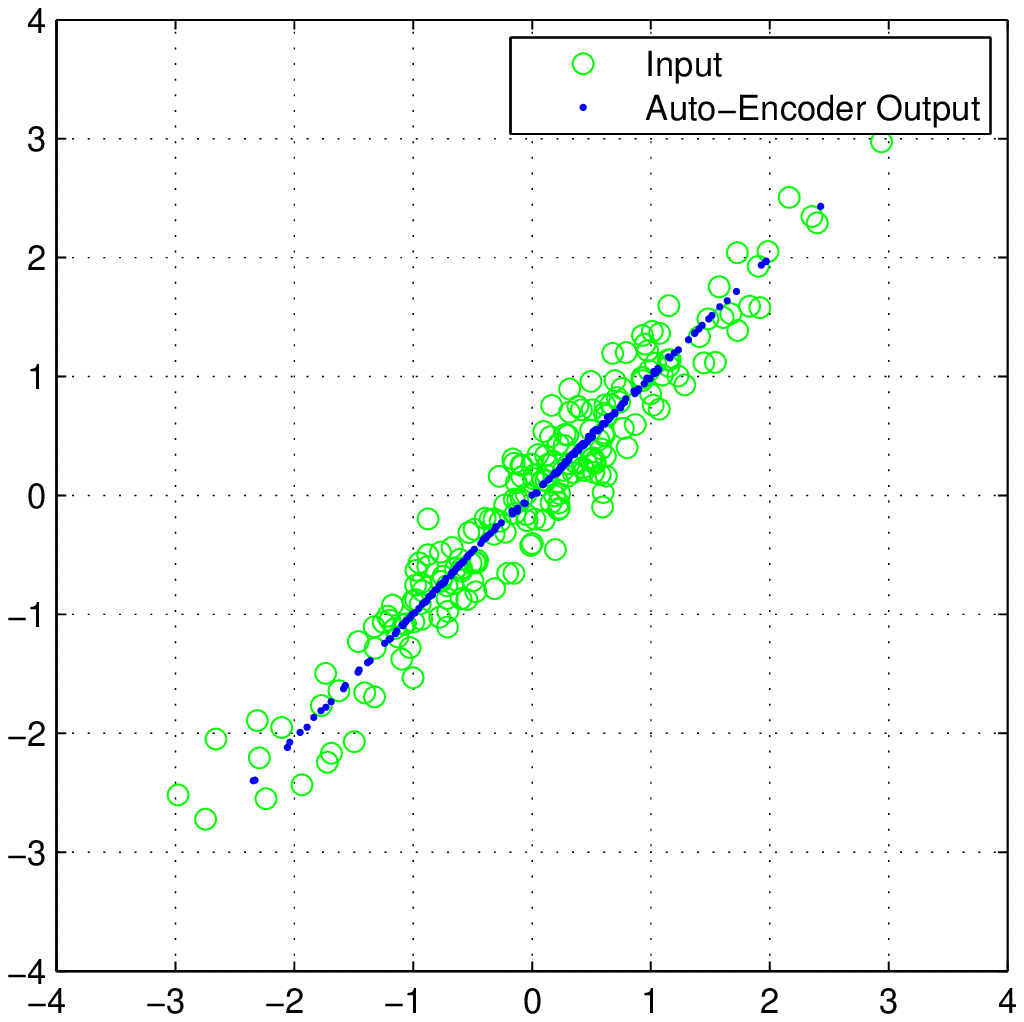}}
\subfigure[$\mathrm{logsig}$ $g(x) = 1/(1+\exp(-x))$]{\includegraphics[width=4.6cm]{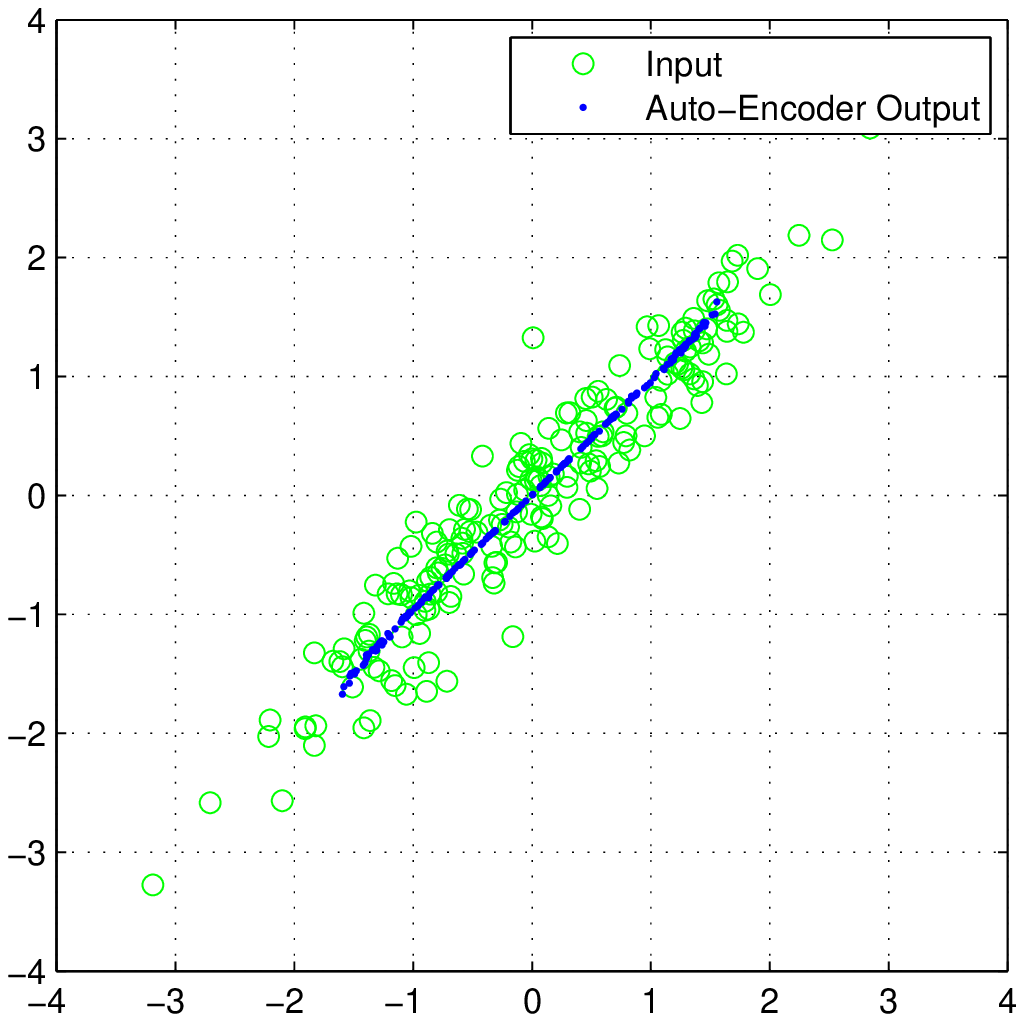}}
\caption{Outputs for different activation functions of the rate-distortion auto-encoder with an over-complete representation when the inputs are Gaussian distributed.}\label{fig:RDAE_Gaussian}
\end{figure} 
 \begin{figure}[!h]
\centering
\subfigure[Linear $g(x) = x$]{\includegraphics[width=6cm]{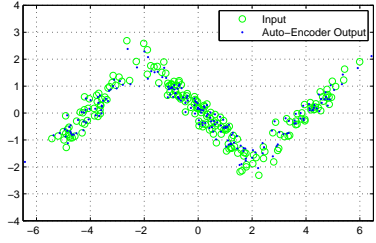}}
\subfigure[ReLU $g(x) = \max\{0,x\}$]{\includegraphics[width=6cm]{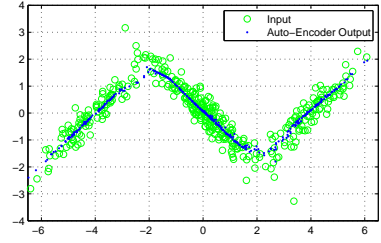}}
\subfigure[$\mathrm{logsig}$ $g(x) = 1/(1+\exp(-x))$]{\includegraphics[width=6cm]{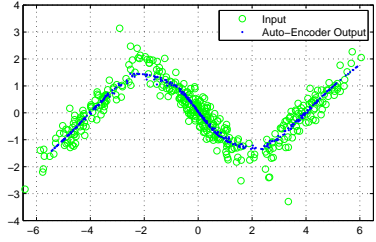}}
\subfigure[SatLU $g(x) = \max\{0,x\} - \max\{0,x-1\}$]{\includegraphics[width=6cm]{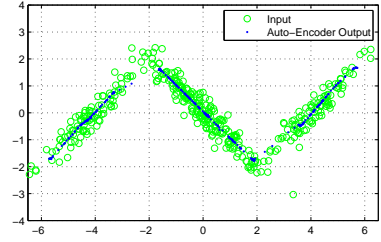}}
\caption{Outputs for different activation functions of the rate-distortion auto-encoder with an over-complete representation when the inputs are a mixture of Gaussian distributions.}\label{fig:RDAE_Gaussian_Mixture}
\end{figure} 
Figure \ref{fig:RDAE_Gaussian} shows the outputs of the three auto-encoders on the Gaussian distributed data. It can be seen that the outputs approximately align with what corresponds roughly to the first principal component of the data. Notice, that no bottleneck neither shrinkage was explicitly defined. The parameters of our cost function are $\mu = 0.5$ for the distortion trade-off,  and $\sigma_x = 0.2\sqrt{2} = \sigma_{\hat{x}}$ for the kernel size.\\ 
\textbf{Mixture of Gaussians:} The second example, employs a mixture of three Gaussian distributions to show the output of the rate-distortion auto-encoder in a nonlinear scenario, where an over-complete representation followed by a nonlinearity can be advantageous. The means of and covariances of the mixture components are
\begin{equation}\begin{split}
\mu_1 = \left(\begin{array}{c} 2 \\-2
\end{array}\right),\:\mu_2 = \left(\begin{array}{c} -2 \\-2
\end{array}\right),\:\mu_1 = \left(\begin{array}{c} 6 \\-2
\end{array}\right);\: \textrm{and} \:\\
\bm{\Sigma}_1 = \left(
\begin{array}{cc}
1 & -0.95 \\
-0.95 & 1
\end{array}
\right),\:\bm{\Sigma}_2 = \bm{\Sigma}_3  = \left(
\begin{array}{cc}
1 & 0.95 \\
0.95 & 1
\end{array}
\right), 
\end{split}
\end{equation}
respectively, and the mixing weights are $p_1 = 0.5$, and $p_2 = p_3= 0.25$.\\
Figure \ref{fig:RDAE_Gaussian_Mixture} shows the outputs of the four auto-encoders on the mixture of Gaussian distributions. The auto-encoders employ: linear, rectified linear, sigmoidal, and saturated linear units. It can be seen that the outputs approximately align with what can be though as the principal curves of the data. Again, we want to stress that no bottleneck neither shrinkage was explicitly defined. In this case each of the auto-encoder has 20 units for encoding, which would easily over fit the data in the absence of any regularization or add-hoc constraints such as tied weights. The parameters of our cost function are $\mu = 0.5$ for the distortion trade-off,  and $\sigma_x = 0.2\sqrt{2} = \sigma_{\hat{x}}$ for the kernel size. The linear units seem to fit the data, but as we previously mentioned they favor the principal components. Lowering the value of $\mu$ would collapse the reconstructed points into a line. This is not necessarily the case when nonlinear units are considered.\\ 
Finally, in Figure \ref{fig:RDAE_Gaussian_Mixture_Energy} we show the resulting energy $\Vert x - \hat{x} \Vert^2$ landscapes for the over-complete auto-encoder with soft rectified linear units after being trained with: Non-regularized, de-noising auto-encoder using isotropic Gaussian noise, and the proposed rate distortion function. 
The rate-distortion objective makes the auto-encoder carve well-defined ravines in the energy landscape at the points were majority of the data lies.
\subsection{Handwritten Digits}
Here we use a subset of 20000 samples from MNIST to train a rate-distortion auto-encoder. Unlike conventional stochastic gradient, that uses mini-batches of randomly sample data to compute the estimate of the gradient at each step, we define mini-batches by pre-clustering the data. The reason behind this procedure is that the Gram matrix employed to compute entropy measure is approximately block diagonal after reordering the samples by clustering. Therefore, the influence of the eigenvectors would be also local around the defined clusters. At first glance, this approach seems to add the computational overhead of the clustering. However, there are cases samples are naturally clustered, for example, sequences of images or processes that are assumed to be piecewise stationary.  
\begin{figure}
  \centering
\subfigure[Analysis]{\includegraphics[height = 7cm]{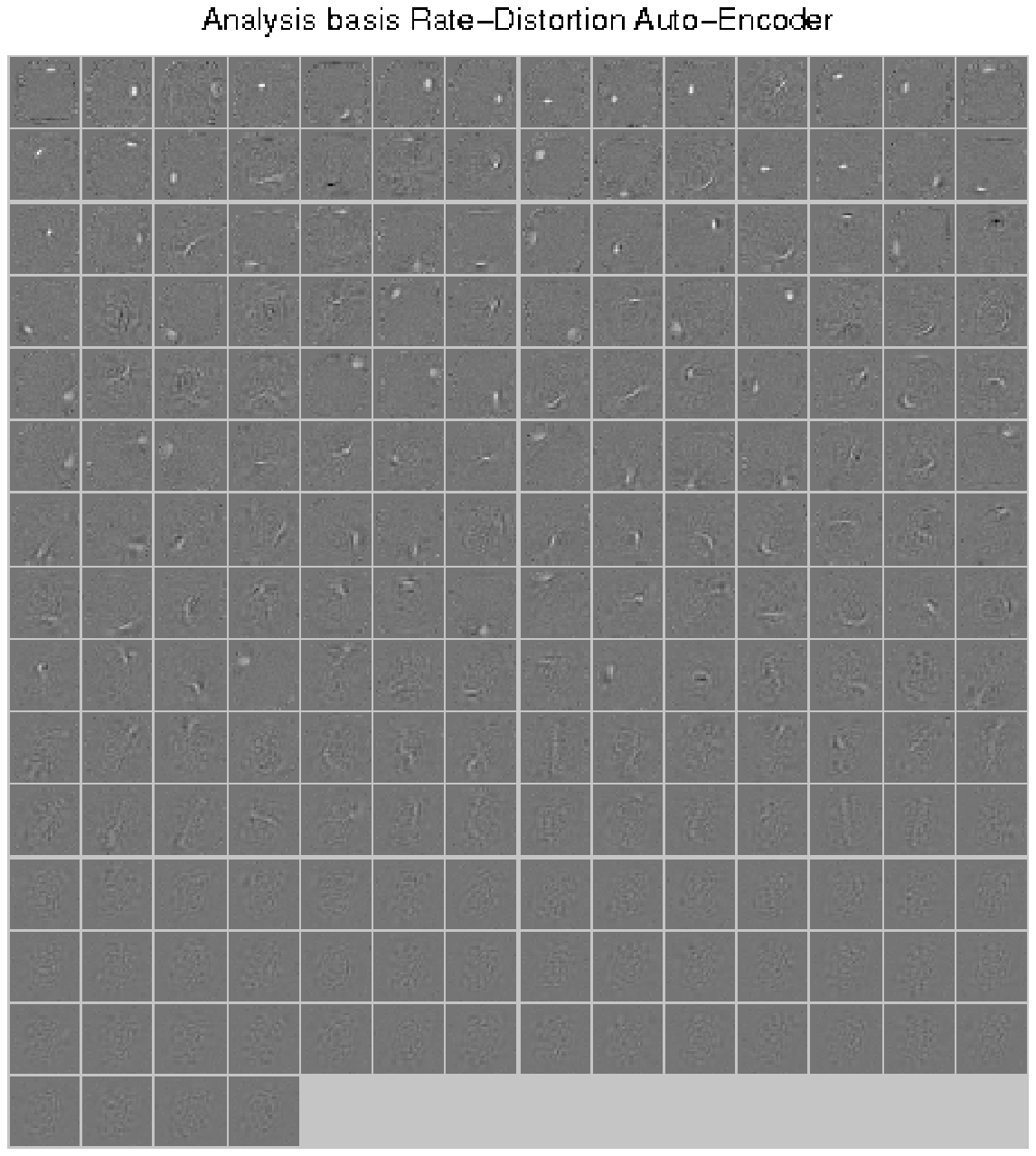}}
\subfigure[Synthesis]{\includegraphics[height = 7cm]{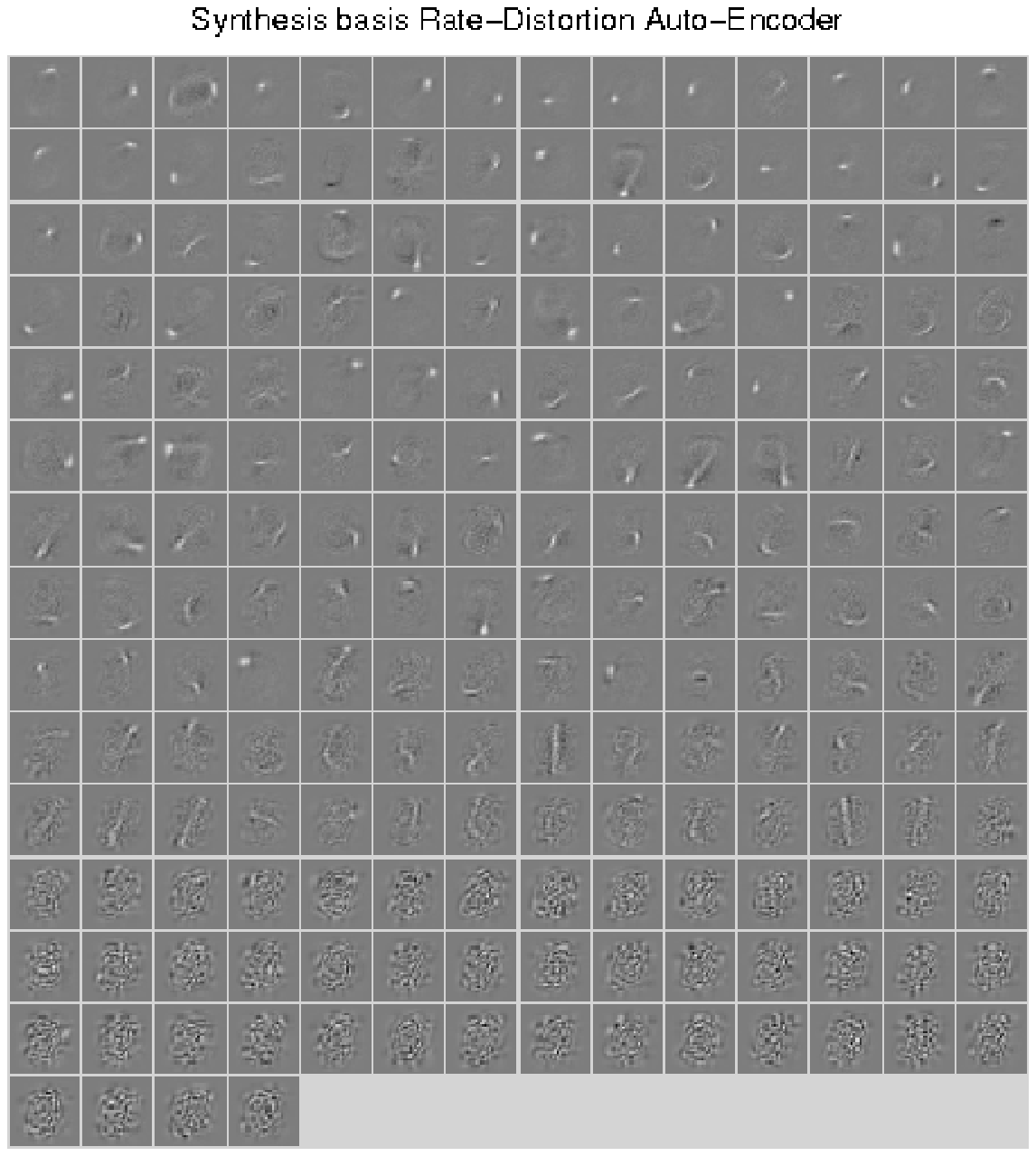}}
\caption{Analysis and Synthesis basis encountered using the Rate-Distortion Auto-Encoder algorithm}\label{fig:RDAE_MNIST}
\end{figure}  
Figure \ref{fig:RDAE_MNIST} shows an example of the learned analysis and synthesis basis, after training an auto-encoder with $200$ $\mathrm{logsig}$ units in the representation layer (hidden layer). The trade off parameter has been set to $0.5$ and the kernel size for the entropy measure is $8.4$. Pixels values were scaled from $o$ to $1$ and then data matrix was centered. In the figure, it can be seen how the auto-encoder not only learn localized blobs as would do normally using tied weights and weight decay, but also learn pen strokes without explicit sparsity constraint or predefined corruption model as presented in the work on d-noising auto-encoders. Varying the trade off parameter gives different regimes for the obtained features. For smaller reconstruction errors features tend to be more localized, whereas for larger tolerance the features go from blobs, to pen strokes, to even full digits. Recall that  No tied-weights were employed. 
\section{Conclusions}
We presented an algorithm for auto-encoders based on a rate-distortion objective that tries to minimize the mutual information between the inputs and outputs subject to a fidelity constraint. As a motivation example, we showed that for multivariate Gaussian distributed data, PCA can be understood as an optimal mapping in the rate-distortion sense. Moreover, we described how the rate-distortion optimization problem can be understood as a learning objective where the fidelity constraint plays the role of a risk functional and the mutual information acts as a regularization term. To provide a training algorithm, we employed a recently introduced measure of entropy based on infinitely divisible matrices that avoids the plug in estimation of densities. Experiments using over-complete bases showed that the auto-encoder was able to learn useful encoding mapping (representation) can learn a regularized input-output in an implicit manner. As future work, we would like to investigate on the relation between the information theoretic objective and the graph Laplacian that arose from the parameter update rules.   
%
\bibliographystyle{IEEEtran}
\bibliography{biblio}

\end{document}